\documentclass[letterpaper]{article} 
\usepackage{aaai2026}  
\usepackage{times}  
\usepackage{helvet}  
\usepackage{courier}  
\usepackage[hyphens]{url}  
\usepackage{graphicx} 
\usepackage{natbib}  
\usepackage{caption} 
\usepackage{booktabs}
\usepackage{amsmath}
\usepackage{amssymb}
\usepackage{xcolor}
\usepackage{subcaption}
\usepackage{algorithm}
\usepackage{algorithmic}

\usepackage{newfloat}
\usepackage{listings}
\DeclareCaptionStyle{ruled}{labelfont=normalfont,labelsep=colon,strut=off} 
\floatstyle{ruled}
\newfloat{listing}{tb}{lst}{}
\floatname{listing}{Listing}
\title{OPLoRA: Orthogonal Projection LoRA Prevents Catastrophic Forgetting during Parameter-Efficient Fine-Tuning}
\author{
    Yifeng Xiong\textsuperscript{\rm 1},
    Xiaohui Xie\textsuperscript{\rm 1}
}
\affiliations{
    \textsuperscript{\rm 1}University of California, Irvine\\

}

\begin{document}

\maketitle

\begin{abstract}
Low-Rank Adaptation (LoRA) enables efficient fine-tuning of large language models but suffers from catastrophic forgetting when learned updates interfere with the dominant singular directions that encode essential pre-trained knowledge. We propose Orthogonal Projection LoRA (OPLoRA), a theoretically grounded approach that prevents this interference through double-sided orthogonal projections. By decomposing frozen weights via SVD, OPLoRA constrains LoRA updates to lie entirely within the orthogonal complement of the top-$k$ singular subspace using projections $P_L = I - U_k U_k^\top$ and $P_R = I - V_k V_k^\top$. We prove that this construction exactly preserves the top-$k$ singular triples, providing mathematical guarantees for knowledge retention. To quantify subspace interference, we introduce $\rho_k$, a metric measuring update alignment with dominant directions. Extensive experiments across commonsense reasoning, mathematics, and code generation demonstrate that OPLoRA significantly reduces forgetting while maintaining competitive task-specific performance on LLaMA-2 7B and Qwen2.5 7B, establishing orthogonal projection as an effective mechanism for knowledge preservation in parameter-efficient fine-tuning.
\end{abstract}


\section{Introduction}

The widespread adoption of large language models has created an urgent need for efficient adaptation techniques that can specialize pre-trained models for downstream tasks without incurring prohibitive computational costs \cite{pmlr-v97-houlsby19a, lester-etal-2021-power, liu2022ptuningv2prompttuning}. Low-Rank Adaptation (LoRA) \cite{lora} has emerged as a leading method for parameter-efficient fine-tuning, enabling task-specific customization by learning low-rank updates to frozen weight matrices while significantly reducing memory and compute requirements.

However, despite its computational advantages, LoRA suffers from a critical limitation: it provides no explicit mechanism to preserve the essential knowledge encoded in pre-trained weights, leading to catastrophic forgetting that degrades general capabilities after task-specific fine-tuning \cite{biderman2024loralearn, yang2025cordacontextorienteddecompositionadaptation, dou2024loramoealleviateworldknowledge}. This forgetting phenomenon presents a substantial challenge for real-world deployment, where models are expected to retain their foundational knowledge while acquiring new, task-specific capabilities. The core issue lies in LoRA's unconstrained optimization process, which permits learned updates to inadvertently interfere with the principal representational subspaces of the pre-trained model.

Our key insight is that essential pre-trained knowledge is concentrated in the dominant singular directions—the subspaces corresponding to the largest singular values of weight matrices. By constraining LoRA updates to lie entirely within the orthogonal complement of these critical subspaces, we can provably prevent interference with pre-trained knowledge while maintaining expressive capacity for task-specific adaptation.

We present Orthogonal Projection LoRA (OPLoRA), which implements this constraint through a simple yet powerful modification: applying orthogonal projections to both sides of the low-rank update. Our method decomposes pre-trained weight matrices using Singular Value Decomposition and constructs projection matrices that remove any component aligned with the top-$k$ singular directions. This ensures that the learned adaptation $\Delta W = P_L B A P_R$ lies entirely in the orthogonal complement of the dominant subspace, where $P_L$ and $P_R$ are orthogonal projectors onto the complement of the top-$k$ left and right singular vectors.

We provide theoretical guarantees proving that this construction exactly preserves the top-$k$ singular triples of the original weight matrices, offering a principled and mathematically grounded solution to the forgetting problem. To quantify subspace interference, we introduce $\rho_k$, a novel metric that measures the proportion of update energy residing within the dominant subspace of pre-trained weights.

Extensive experiments across commonsense reasoning, mathematics, and code generation on LLaMA-2 7B \cite{llama2} and Qwen2.5 7B \cite{qwen2.5} demonstrate that OPLoRA reduces catastrophic forgetting while maintaining competitive task-specific performance. Our approach consistently surpasses existing LoRA variants in retaining knowledge across diverse benchmarks.

We summarize our contributions as follows:
\begin{itemize}
    \item We present an analysis of catastrophic forgetting in LoRA by introducing $\rho_k$, a subspace alignment metric that quantitatively measures interference with pre-trained knowledge.
    \item We propose Orthogonal Projection LoRA, a theoretically grounded method that provably preserves the dominant singular directions of pre-trained weights through double-sided orthogonal projections.
    \item We validate our approach through extensive experiments across three domains and two model architectures, demonstrating significant improvements in forgetting resistance while maintaining competitive adaptation performance.
\end{itemize}

\section{Related Works}

\subsection{Low-Rank Adaptation and Extensions}
LoRA \cite{lora} is a foundational method for parameter-efficient fine-tuning (PEFT) that introduces trainable low-rank matrices into each weight layer while keeping the pre-trained weights frozen. The core idea is that task-specific knowledge can be captured through a low-dimensional subspace, reducing both memory and computational requirements. Recent work has extended this framework through sophisticated weight decomposition strategies, notably DoRA \cite{dora}, which addresses the limited expressiveness of vanilla LoRA by decomposing pre-trained weight matrices into magnitude and direction components and applying low-rank adaptation exclusively to the direction component. A parallel line of research has focused on exploiting the inherent structure of pre-trained weights through singular value decomposition, with PiSSA \cite{pissa} replacing LoRA's random initialization with principled construction using the principal singular vectors and values of frozen weight matrices, enabling direct modification of the essential subspace and improved alignment between low-rank updates and full fine-tuning behavior. In contrast, MiLoRA \cite{milora} adopts a complementary approach by explicitly preserving the dominant subspace of pre-trained weights, performing SVD on each weight matrix and freezing the principal components during fine-tuning while updating only the minor singular components via low-rank adaptation, thereby constraining initialization to a subspace orthogonal to core pre-trained knowledge and minimizing interference with essential representations. Additionally, OLoRA \cite{olora} focuses on improving optimization landscapes by initializing adaptation matrices with orthonormal bases derived through QR decomposition, constraining updates to well-conditioned subspaces to promote improved gradient flow and convergence stability across diverse tasks. 

\subsection{Catastrophic Forgetting and Knowledge Preservation}
Catastrophic forgetting represents a fundamental challenge in continual learning and fine-tuning, where models lose previously acquired knowledge when learning new tasks \cite{ramasesh2020anatomy}. In the context of large language models, this phenomenon manifests as degraded performance on general capabilities after task-specific fine-tuning, posing significant practical limitations for real-world deployment. The measurement and mitigation of catastrophic forgetting have become critical research directions, with various approaches emerging to quantify and address this problem.
Early mitigation strategies focused on regularization and replay mechanisms. Elastic Weight Consolidation (EWC) \cite{ewc} constrains training updates with a Fisher information-weighted quadratic penalty to identify and protect crucial weights from previous tasks. Learning Without Forgetting (LwF) \cite{lwf} addresses forgetting without retaining old data by recording the original network's soft outputs on new-task data, then training the model to match those pseudo-labels while learning new labels. Meta-Experience Replay (MER) \cite{mer} combines each incoming example with a small reservoir of past samples and applies lightweight Reptile meta-updates that encourage gradient alignment across old and new data, maximizing transfer while reducing interference. Orthogonal Gradient Descent (OGD) \cite{ogd} tackles catastrophic forgetting by constraining every new gradient step to be orthogonal to the gradient directions that mattered for earlier tasks. Instead of storing old data, it stores a compact set of per‑task prediction‑gradient vectors; before each parameter update, the current task’s gradient is projected into the null‑space of those stored directions.
Recent research has increasingly explored parameter-efficient fine-tuning techniques as a natural solution to knowledge preservation. Within the LoRA framework, MiLoRA attempts to preserve dominant pre-trained weights by initializing LoRA matrices with minor singular components, constraining adaptation to less critical subspaces. However, MiLoRA does not explicitly address the problem of catastrophic forgetting. The most recent work, LoRA-Null \cite{loranull}, mitigates catastrophic forgetting by leveraging the null space for adapter initialization and constructing LoRA adapters from projections of pre-trained weights onto the null space of representative activations to ensure that update directions do not interfere with pre-trained knowledge.

\begin{figure*}[t]
    \centering
    \includegraphics[width=0.9\textwidth]{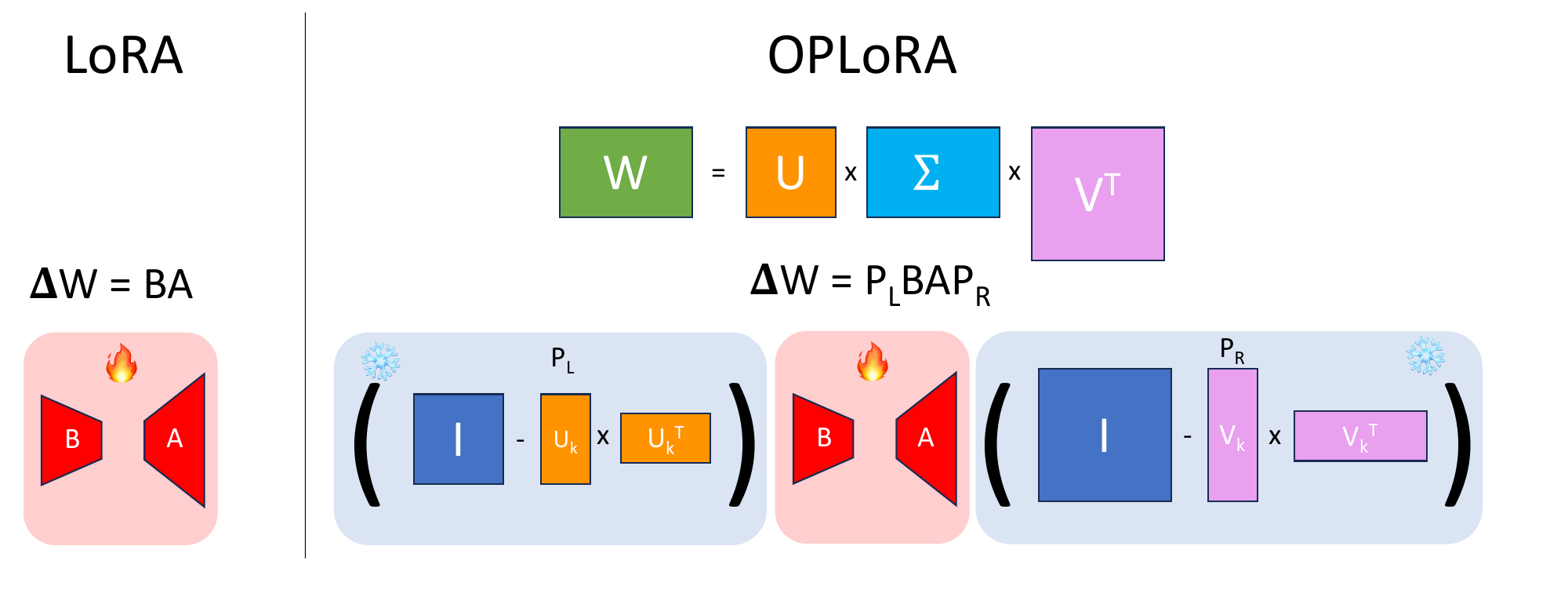}
    \caption{
    Comparison of standard LoRA (left) and our proposed OPLoRA (right).
    The pre-trained weight matrix $W$ is decomposed via singular value decomposition (SVD) as $W = U \Sigma V^\top$.
    During fine-tuning, the standard LoRA update $BA$ is sandwiched between two frozen projection matrices,
    $P_L = (I - U_k U_k^\top)$ and $P_R = (I - V_k V_k^\top)$.
    This structure constrains the low-rank adaptation to lie in the orthogonal complement of the pre-trained dominant subspace,
    mitigating interference with essential model knowledge and reducing catastrophic forgetting.
    }
    \label{fig:initialization}
\end{figure*}

\section{Methodology}

\subsection{Preliminaries: Low-Rank Adaptation (LoRA)}

Consider a pre-trained weight matrix $W_0 \in \mathbb{R}^{d_{\text{out}} \times d_{\text{in}}}$. LoRA proposes that task-specific adaptation can be achieved by learning a low-rank update $\Delta W = \frac{\alpha}{r} BA$ instead of updating the full matrix. The forward pass becomes:
\begin{equation}
h = W_0 x + \Delta W x = W_0 x + \frac{\alpha}{r} BAx,
\end{equation}
where $B \in \mathbb{R}^{d_{\text{out}} \times r}$ and $A \in \mathbb{R}^{r \times d_{\text{in}}}$ are trainable matrices, and $r \ll \min(d_{\text{out}}, d_{\text{in}})$ is the rank. The scaling factor $\alpha$ ensures the update is appropriately scaled.

During fine-tuning, $W_0$ remains frozen while gradients only flow through $A$ and $B$. This strategy significantly reduces the number of trainable parameters and enables efficient adaptation, particularly for large models. LoRA normally preserves the pre-trained model’s behavior by initializing 
$A$ with Kaiming-uniform \cite{he2015delving} values and setting $B=0$.

\subsection{Understanding and Quantifying Forgetting in LoRA}
\label{sec:subspace_analysis}

While LoRA effectively reduces parameter count, it does not explicitly account for the distribution of information in $W_0$. In practice, fine-tuning with LoRA can lead to catastrophic forgetting, where essential knowledge encoded in the pre-trained model is overwritten during adaptation fine-tuning.

The forgetting phenomenon is often implicit and hard to quantify. To better understand it, we analyze the learned adaptation $\Delta W$ relative to the singular directions of $W_0$. Specifically, we ask: Does the LoRA update interfere with the most important components of the $W_0$?

To answer this, we leverage the Singular Value Decomposition, which decomposes the weight matrix $W_0$ into:
\begin{equation}
W_0 = U \Sigma V^\top,
\end{equation}
where $U$ and $V$ are orthogonal matrices, and $\Sigma$ is a diagonal matrix of singular values sorted in descending order. The leading $k$ left singular vectors, $U_k \in \mathbb{R}^{d_{\text{out}} \times k}$, define the principal column subspace of $W_0$, embodying the directions with maximal spectral energy or task-critical information.

Empirical observations \cite{li2018measuringintrinsicdimensionobjective,aghajanyan2020intrinsic, lora} show that singular values of $W_0$ decay rapidly, meaning the essential semantics of the layer are concentrated in a low-dimensional subspace. However, since the LoRA weight is unconstrained during training, the resulting update $\Delta W$ may interfere with these top directions—modifying them and causing loss of important pre-trained knowledge.

To make this interference measurable, we define a subspace alignment metric $\rho_k$, which quantifies how much of $\Delta W$ lies within the top-$k$ singular directions of $W_0$.

Let $Q_k = U_k U_k^\top$ denote the orthogonal projector onto the column space of $W$. Then, we can decompose the LoRA update $\Delta W$ as:
\begin{equation}
\Delta W = Q_k \Delta W + (I - Q_k) \Delta W,
\end{equation}
where the first term lies within the column space of $W$, and the second term lies in its orthogonal complement. Using the Pythagorean property of orthogonal projections, we have:
\begin{equation}
    ||\Delta W||_F^2 = ||Q_k \Delta W||_F^2 + ||(I - Q_k) \Delta W||_F^2.
\end{equation}
Then we define:
\begin{equation}
    \rho_k = \frac{||Q_k \Delta W||_F^2}{||\Delta W||_F^2} \in [0, 1],
\end{equation}
which measures the proportion of the updated spectral energy that lies within the original top-k column space of $W$.

A value of $\rho_k \rightarrow 1$ implies that the LoRA update stays mostly in the original subspace and may risk interfering with pre-trained features. Conversely, $\rho_k \rightarrow 0$ indicates that the update explores directions orthogonal to the pre-trained space—potentially enabling new task-specific representations while preserving prior knowledge.

\subsection{Preserving Pre-trained Knowledge via Orthogonal Projection}
To avoid disrupting the dominant directions of $W_0$, we confine the LoRA update to the orthogonal complement of its top-$k$ subspace.
Let $U_k \in \mathbb{R}^{d_{\text{out}} \times k}$ be the matrix of the top-$k$ left singular vectors of $W_0$ ($k$ is a hyperparameter). Define the orthogonal projector onto the complement of $\text{span}\{U_k\}$
\begin{equation}
P = I - U_k U_k^\top.
\end{equation}
We will show that the LoRA update $\Delta W=PBA$ is entirely confined to this complementary subspace, ensuring it contains no components along the dominant $k$ singular directions.

\paragraph{Proposition 1.}
Let $U_k \in \mathbb{R}^{d_{\text{out}} \times k}$ be an orthonormal basis for the top-$k$ left singular vectors of  $W_0 \in \mathbb{R}^{d_{\text{out}} \times d_{\text{in}}}$. Define the orthogonal projector onto the complement of $\text{span}\{ U_k\}$ as $P = I - U_k U_k^\top$. For every LoRA update  $\Delta W = PBA$ and any input vector $x$, the output $\Delta W x = PBAx$ resides completely in the orthogonal complement of $\text{span}\{U_k\}$, i.e.,
\begin{equation}
U_k^\top (PBAx) = 0 \quad \text{for all } x.
\end{equation}

\paragraph{Proof.}
We first verify that $P$ is a projection matrix: $P^2 = (I - U_k U_k^\top)^2 = I - 2U_k U_k^\top + U_k U_k^\top U_k U_k^\top = I - U_k U_k^\top = P,$ since $U_k^\top U_k = I$.

Now for any input $x \in \mathbb{R}^{d_{\text{in}}}$, the LoRA output is $\Delta W x = PBAx.$
We show that this vector is orthogonal to $\text{span}(U_k)$ by checking:
\begin{align*}
    U_k^\top (PBAx) &= U_k^\top (I - U_k U_k^\top) BAx \\
             &= (U_k^\top - U_k^\top U_k U_k^\top) BAx \\
             &= (U_k^\top - U_k^\top) BAx \\
             &= 0.
\end{align*}

Hence, $\Delta W x \in \text{span}(U_k)^\perp$, and the LoRA update does not interfere with the dominant subspace of the original weight matrix.
\hfill$\blacksquare$

While Proposition 1 guarantees that the LoRA update $\Delta W = PBA$ does not inject any output into the top-$k$ left singular directions of $W_0$, it is natural to ask whether any stronger preservation property holds. In particular, does the use of the projection $P = I - U_k U_k^\top$ preserve the original singular vectors $u_i$ or the singular values $\sigma_i$?

The answer is no. Even though $\Delta W x$ is orthogonal to $\text{span}(U_k)$ for every $x$, this alone does not guarantee that adding $\Delta W$ preserves the top-$k$ singular vectors or singular values of the updated weight matrix $W' = W_0 + \Delta W$. To see this, assume for contradiction that $W' u_i = \sigma_i v_i$ for some $i \le k$. Then we would imply
\begin{equation}
\Delta W u_i = W' u_i - W_0 u_i = \sigma_i v_i - \sigma_i v_i = 0.
\end{equation}
However, this conclusion does not follow from Proposition 1, because $\Delta W u_i$ need not be zero—it only lies in $\text{span}(U_k)^\perp$. Hence, the top-$k$ left singular vectors of $W_0$ are generally not preserved.

This leads to a stronger goal: can we preserve each complete singular triple $(u_i, \sigma_i, v_i)$? We can — by projecting on {\bf both} sides. 
Constraining $\Delta W$ to the orthogonal complements of the top-$k$ left and right singular subspaces guarantees that the top-
$k$ singular vectors and singular values of
$W_0$ remain intact. We now state and prove this claim formally.

\paragraph{Proposition 2.}
Let $W_0 \in \mathbb{R}^{d_{\text{out}} \times d_{\text{in}}}$ admit the singular value decomposition
\begin{equation}
W_0 = U \Sigma V^\top = U_k \Sigma_k V_k^\top + U_\perp \Sigma_\perp V_\perp^\top,
\end{equation}
where $U_k = [u_1, \dots, u_k]$, $V_k = [v_1, \dots, v_k]$ collect the top-$k$ left and right singular vectors,  $\Sigma_k = \operatorname{diag}(\sigma_1, \dots, \sigma_k)$ with $\sigma_1 \ge \dots \ge \sigma_k > 0$, and $U_\perp$,$\Sigma_\perp$, and  $V_\perp$ contain remaining singular triples.
Define the orthogonal projectors on the complements of the dominant subspaces:
\begin{equation}
P_L = I - U_k U_k^\top, \quad P_R = I - V_k V_k^\top.
\end{equation}
Construct the LoRA update as $\Delta W = P_L B A P_R$ for arbitrary matrices $A$ and $B$, and  set updated weight matrix $W' = W_0 + \Delta W$. Then, for every $i = 1, \dots, k$, 
\begin{equation}
W' v_i = \sigma_i u_i, \quad (W')^\top u_i = \sigma_i v_i,
\end{equation}
or equivalently,
\begin{equation}
U_k^\top W' V_k = \Sigma_k.
\end{equation}
Thus the top-$k$ singular vectors and their corresponding singular values of $W_0$ remain exactly preserved after the update.

\paragraph{Proof.}

By the definitions of $P_L$ and $P_R$, we have:
\begin{equation}
P_L U_k = 0, \quad U_k^\top P_L = 0, \quad P_R V_k = 0, \quad V_k^\top P_R = 0.
\end{equation}

Applying the identities above:
\begin{equation}
\Delta W V_k = P_L B A (P_R V_k) = P_L B A \cdot 0 = 0,
\end{equation}
\begin{equation}
\Delta W^\top U_k = P_R A^\top B^\top (P_L U_k) = P_R A^\top B^\top \cdot 0 = 0,
\end{equation}

For each $i \le k$, since $\Delta W v_i = 0$, we have:
\begin{equation}
W' v_i = (W_0 + \Delta W) v_i = W_0 v_i = \sigma_i u_i.
\end{equation}
Similarly, since $\Delta W^\top u_i = 0$:
\begin{equation}
(W')^\top u_i = W_0^\top u_i = \sigma_i v_i.
\end{equation}
Thus, $(u_i, \sigma_i, v_i)$ remains an exact singular triple of $W'$.

Equivalently we have:
\begin{equation}
U_k^\top W' V_k = U_k^\top W_0 V_k = \Sigma_k.
\end{equation}
\hfill$\blacksquare$

\paragraph{Our Method.}
Motivated by Proposition 2, we propose our one-line modification to LoRA that constrains the update $\Delta W$ to lie entirely in the orthogonal complement of the top-$k$ singular subspace of the pre-trained weight matrix $W_0$. Specifically, we apply projections on both sides of the low-rank update and define:
\begin{equation}
\Delta W = P_L B A P_R,
\end{equation}
where $P_L = I - U_k U_k^\top$ and $P_R = I - V_k V_k^\top$ are orthogonal projectors that remove any component aligned with the top-$k$ left and right singular vectors of $W_0$, respectively. This construction guarantees that the most informative singular directions of $W_0$ are exactly preserved under the update, thus mitigating forgetting while still allowing for expressive task-specific adaptation in the residual subspace.

\begin{figure*}[t]
    \centering
    \begin{subfigure}[t]{0.45\linewidth}
        \centering
        \includegraphics[width=\linewidth]{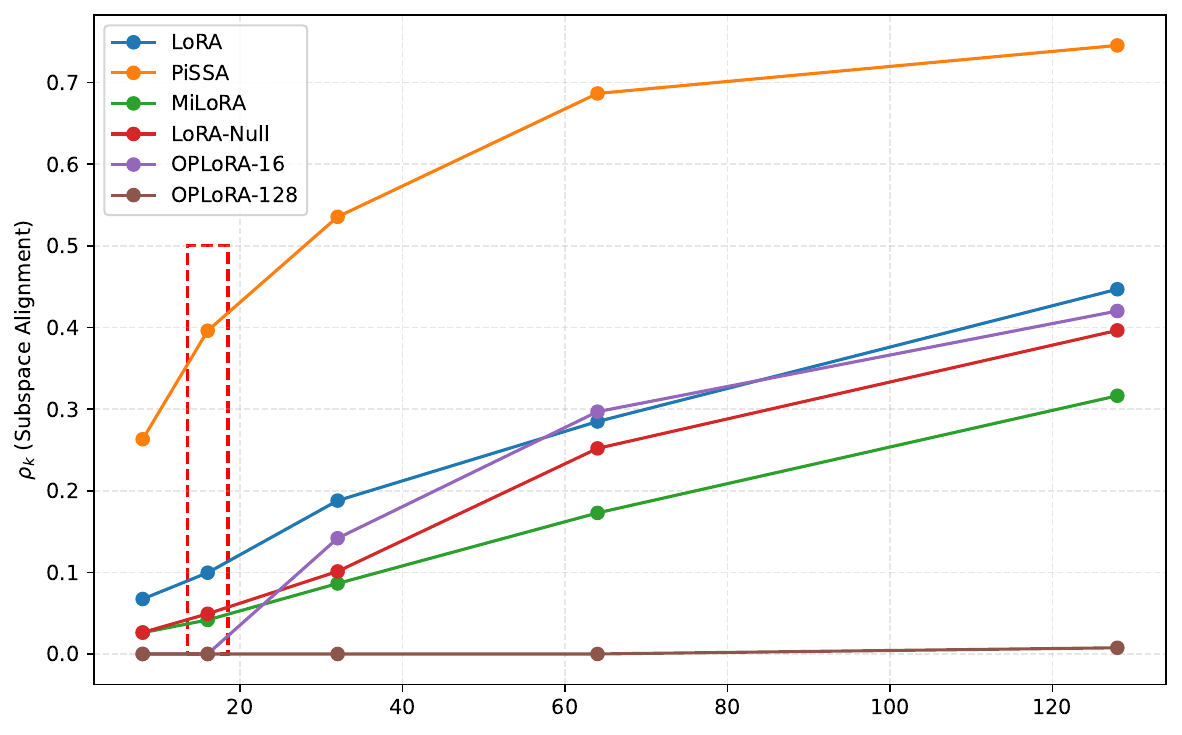}
        \caption{Subspace alignment with top-$k$ directions.}
        \label{fig:metamath_rho_llama}
    \end{subfigure}
    \hfill
    \begin{subfigure}[t]{0.45\linewidth}
        \centering
        \includegraphics[width=\linewidth]{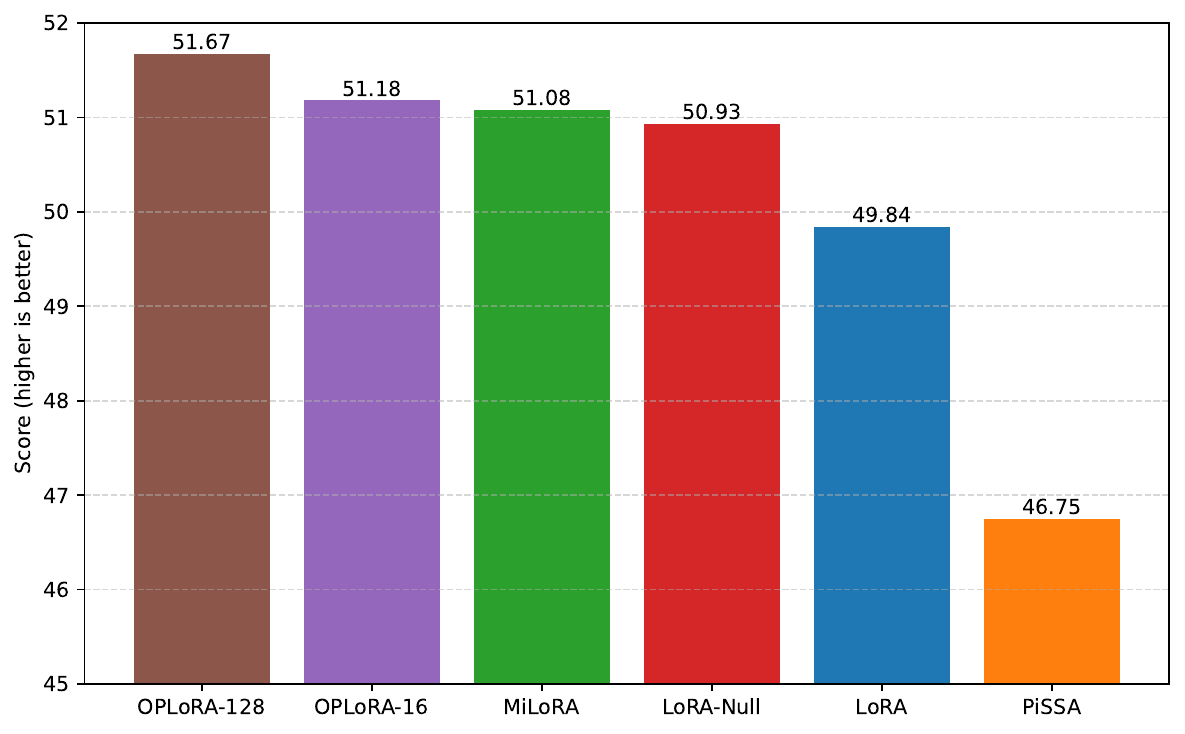}
        \caption{Comparison of Average Forgetting Evaluation Score.}
        \label{fig:metamath_bar_llama}
    \end{subfigure}
    \caption{
    Quantifying and Mitigating Forgetting in LoRA-based Fine-Tuning.
    \textbf{(a)} Subspace alignment $\rho_k$ between the learned LoRA update and the top-$k$ singular directions of the pre-trained weight matrix $W_0$ in the LLaMA-2 7B \texttt{layers.0.self.attn.q\_proj}. A higher $\rho_k$ indicates stronger interference with pre-trained knowledge. OPLoRA-128 achieves the lowest alignment across all $k$, indicating better preservation of prior knowledge. Note that $\rho_k$ is not exactly zero for OPLoRA-128 due to the use of a low-rank SVD approximation (\texttt{svd\_lowrank}) for computational efficiency.
    \textbf{(b)} Average forgetting scores evaluated on three held-out commonsense reasoning tasks after fine-tuning. Higher scores indicate better retention of pre-trained knowledge. Methods are sorted by their corresponding subspace alignment value $\rho_{16}$ (shown in (a)). A clear negative correlation is observed: lower $\rho_{16}$ aligns with better forgetting robustness.
    }
    \label{fig:forgetting_analysis}
\end{figure*}

\section{Experiments}

 We conduct comprehensive experiments to evaluate the effectiveness of OPLoRA in preventing catastrophic forgetting while maintaining competitive fine-tuning performance. Our comparisons include four LoRA-based baselines: LoRA \cite{lora}, PiSSA \cite{pissa}, MiLoRA \cite{milora}, and LoRA-Null \cite{loranull}. Each method is evaluated across three task domains, commonsense reasoning, mathematics, and Python code generation, using two backbone models: LLaMA-2 7B \cite{llama2} and Qwen2.5 7B \cite{qwen2.5}. All methods are trained under the same hyperparameter configuration for fair comparison, with LoRA applied to the self-attention and MLP layers, specifically targeting the modules $\mathtt{q\_proj}, \mathtt{v\_proj}, \mathtt{up\_proj}, \mathtt{down\_proj}, \mathtt{o\_proj}$. 
 For OPLoRA, we fix the projection rank $k$ to two values, $k=16$ and $k=128$, across all experiments. These ranks are not tuned for performance and are chosen to represent low and high subspace preservation regimes. Exploration of optimal projection rank selection is left for future work.

\begin{table}[t]
\centering
\small
\begin{tabular}{lcc}
\toprule
\textbf{Method} & \textbf{SVD Without $W_{\text{res}}$} & \textbf{Forgetting Mitigation} \\
\midrule
LoRA        & $\times$     & $\times$ \\
PiSSA       & $\times$     & $\times$ \\
MiLoRA      & $\times$    & $\times$ \\
LoRA-Null   & $\times$     & $\checkmark$ \\
OPLoRA      & $\checkmark$ & $\checkmark$ \\
\bottomrule
\end{tabular}
\caption{Comparison of LoRA-based methods on whether they utilize SVD-based initialization without requiring the residual weight matrix \(W_{\text{res}}\), and whether they explicitly address catastrophic forgetting.}
\label{tab:forgetting_wres}
\end{table}


\subsection{Commonsense Reasoning}
The commonsense reasoning benchmark consists of eight sub-tasks, BoolQ \cite{boolq}, PIQA \cite{piqa}, SIQA \cite{siqa}, HellaSwag \cite{hellaswag}, WinoGrande \cite{winogrande}, ARC-e, ARC-c \cite{arc}, and OBQA \cite{obqa}. Models are fine-tuned on Commonsense170k \cite{hu2023llm} with evaluation performed on individual task test sets using accuracy as the primary metric. Results are presented in Tables~\ref{tab:commonsense_llama} and \ref{tab:commonsense_qwen}.
 
OPLoRA demonstrates exceptional adaptation performance across both model architectures. On LLaMA-2 7B, OPLoRA-128 secures the highest accuracy on four tasks: BoolQ (82.78\%), HellaSwag (75.38\%), ARC-c (52.39\%), and OBQA (45.8\%). Meanwhile, OPLoRA-16 consistently achieves second-best performance on BoolQ, PIQA, HellaSwag, and ARC-c. This pattern extends to Qwen2.5 7B, where OPLoRA-16 leads on four tasks (HellaSwag, WinoGrande, ARC-e, OBQA) while OPLoRA-128 excels on PIQA and ARC-c. The consistent top-two ranking across diverse reasoning challenges demonstrates that our orthogonal projection approach preserves adaptation quality across different projection ranks.

To assess catastrophic forgetting, we evaluate retention of general knowledge on three held-out domains: MathQA (mathematical reasoning) \cite{MathQA}, MBPP (code generation) \cite{MBPP}, and RACE (reading comprehension) \cite{RACE}. These cross-domain evaluations provide critical insights into knowledge preservation during fine-tuning. As shown on the right side of Table~\ref{tab:commonsense_llama}, OPLoRA exhibits strong resistance to catastrophic forgetting on LLaMA-2 7B. Among all methods, OPLoRA-128 achieves the highest accuracy across all three held-out tasks, closely followed by OPLoRA-16. These results highlight the effectiveness of preserving the dominant subspace of pre-trained weights in maintaining general knowledge during fine-tuning. Similar improvements are observed with Qwen2.5 7B. OPLoRA-16 achieves the highest accuracy on MBPP and ranks second on both MathQA and RACE, while OPLoRA-128 obtains the best performance on MathQA and the second-best on MBPP.

\begin{table*}[t]
\centering
\small
\setlength{\tabcolsep}{4pt}
\begin{tabular}{lcccccccc|ccc|c}
\toprule
\multicolumn{1}{c}{} & \multicolumn{8}{c|}{\textbf{Fine-Tuning Evaluation}} & \multicolumn{3}{c|}{\textbf{Forgetting Evaluation}}  \\
\cmidrule(lr){2-9} \cmidrule(lr){10-12} 
\textbf{Method} & \textbf{BoolQ} & \textbf{PIQA} & \textbf{SIQA} & \textbf{HSwag} & \textbf{WinoG} & \textbf{ARC-e} & \textbf{ARC-c} & \textbf{OBQA} 
& \textbf{MathQA} & \textbf{MBPP} & \textbf{RACE} & \textbf{Avg} \\
\midrule
\multicolumn{13}{l}{\textit{Baseline Methods}} \\
LoRA & 81.00 & \textbf{80.19} & 51.58 & 74.73 & 72.84 & 79.29 & 50.51 & 44.8 & \underline{29.08} & \underline{20.2} & \underline{43.06} & 57.03 \\
PiSSA & 62.11 & 55.11 & 34.80 & 28.33 & 49.32 & 28.45 & 26.79 & 26.4 & 22.88 & 0 & 22.39 & 32.42 \\
MiLoRA & 82.04 & 78.99 & \textbf{54.24} & 75.04 & \textbf{73.40} & 79.08 & 52.21 & \underline{45.4} & 28.21 & 18.6 & 41.4 & 57.15 \\
LoRA-Null & 82.02 & 77.97 & 51.43 & 74.74 & 72.88 & 77.48 & 48.29 & 42.6 & 28.98 & 9.8 & 42.20 & 55.31\\
\midrule
\multicolumn{13}{l}{\textit{Our Methods}} \\
OPLoRA-16 & \underline{82.42} & \underline{79.16} & 52.92 & \underline{75.35} & 72.30 & \textbf{79.55} & \underline{52.30} & 45.2 & \underline{29.08} & \underline{20.2} & \underline{43.06} & \underline{57.41}\\
OPLoRA-128 & \textbf{82.78}& 79.11 & \underline{53.53} & \textbf{75.38} & \underline{73.24} & \underline{79.25} & \textbf{52.39} & \textbf{45.8} & \textbf{29.88} & \textbf{20.4} & \textbf{43.63} & \textbf{57.76}\\
\bottomrule
\end{tabular}
\caption{Comparison of LoRA-based fine-tuning methods on commonsense reasoning and forgetting benchmarks. We evaluate each method on eight commonsense tasks (left) to assess fine-tuning performance, and on three held-out tasks (right) to measure the extent of catastrophic forgetting. All models are fine-tuned from LLaMA-2 7B. Bold and underlined values indicate the best and second-best scores.}
\label{tab:commonsense_llama}
\end{table*}

\begin{table*}[t]
\centering
\small
\setlength{\tabcolsep}{4pt}
\begin{tabular}{lcccccccc|ccc|c}
\toprule
\multicolumn{1}{c}{} & \multicolumn{8}{c|}{\textbf{Fine-Tuning Evaluation}} & \multicolumn{3}{c|}{\textbf{Forgetting Evaluation}}  \\
\cmidrule(lr){2-9} \cmidrule(lr){10-12} 
\textbf{Method} & \textbf{BoolQ} & \textbf{PIQA} & \textbf{SIQA} & \textbf{HSwag} & \textbf{WinoG} & \textbf{ARC-e} & \textbf{ARC-c} & \textbf{OBQA} 
& \textbf{MathQA} & \textbf{MBPP} & \textbf{RACE} & \textbf{Avg} \\
\midrule
\multicolumn{13}{l}{\textit{Baseline Methods}} \\
LoRA & \textbf{86.75} & \textbf{79.76} & 51.43 & \underline{78.12} & 73.95 & \underline{76.26} & 55.03 & 45.8 & 50.35 & 58.6 & 39.62 & 63.24 \\
PiSSA & 75.04 & 75.08 & 52.04 & 71.75 & 75.92 & 69.23 & 43.85 & 42.0 & 31.86 & 0.6 & \textbf{40.57} & 52.54 \\
MiLoRA & 86.05 & 77.36 & \textbf{53.42} & 74.73 & \underline{76.47} & 71.42 & 52.13 & 44.4 & 46.47 & 2.6 & 39.62 & 56.79\\
LoRA-Null & 83.30 & 77.61 & 51.57 & 75.67 & 74.51 & 75.84 & 54.83 & 46.8 & 50.24 & 19.8 & 40.11 & 59.12\\
\midrule
\multicolumn{13}{l}{\textit{Our Methods}} \\
OPLoRA-16 & \underline{86.64} & \underline{79.38} & \underline{52.56} & \textbf{79.01} & \textbf{76.64} & \textbf{78.54} & \underline{55.46} & \textbf{47.8} & \underline{50.92} & \textbf{59.6} & \underline{40.28} & \textbf{64.26}\\
OPLoRA-128 & \underline{86.64} & \textbf{79.76} & 51.43 & 77.62 & 74.03 & 76.14 & \textbf{56.14} & \underline{47.4} & \textbf{51.29} & \underline{59.2} & 39.23 & \underline{63.53}\\

\bottomrule
\end{tabular}
\caption{Comparison of LoRA-based fine-tuning methods on commonsense reasoning and forgetting benchmarks. All models are fine-tuned from Qwen2.5 7B. Bold and underlined values indicate the best and second-best scores.}
\label{tab:commonsense_qwen}
\end{table*}

\subsection{Mathematical Reasoning}

For mathematical reasoning, we fine-tune each model on the first 100K examples from the MetaMathQA dataset \cite{metamath}. Evaluation is performed on two in-domain tasks: MATH \cite{math} and GSM8K \cite{gsm8k}, and we report Exact Match (EM) against the ground truth for each test set. As shown in Table~\ref{tab:metamath_llama} and Table~\ref{tab:metamath_qwen}, OPLoRA achieves strong adaptation performance compared to baseline methods on both LLaMA-2 7B and Qwen2.5 7B.

On LLaMA-2 7B, OPLoRA-16 achieves the best performance on both MATH (7.04\%) and GSM8K (49.73\%), with OPLoRA-128 closely following with 6.88\% and 49.35\%. Both variants outperform all baseline methods. For Qwen2.5 7B, a similar pattern emerges. OPLoRA-16 achieves the highest score on both MATH (47.34\%) and GSM8K (83.70\%), while OPLoRA-128 is competitive with 46.72\% and matches the top GSM8K score (83.70\%).

To assess catastrophic forgetting, we evaluate each model on three held-out commonsense reasoning tasks: ARC-e, ARC-c, and SIQA. OPLoRA-128 obtains the best overall forgetting performance on LLaMA-2 7B, with OPLoRA-16 following closely behind. On Qwen2.5 7B, OPLoRA-128 again delivers the strongest forgetting resistance, achieving the best scores on ARC-e and SIQA, and performing competitively on ARC-c. OPLoRA-16 achieves the top score on ARC-c and ranks second on ARC-e and SIQA. Compared to all baselines, both OPLoRA variants show more balanced retention across all three tasks.

\begin{table}[t]
\centering
\small
\setlength{\tabcolsep}{2pt}
\begin{tabular}{lcc|ccc|c}
\toprule
\multicolumn{1}{c}{} & \multicolumn{2}{c}{\textbf{Fine-Tuning Eval}} & \multicolumn{3}{c}{\textbf{Forgetting Eval}} & \\
\cmidrule(lr){2-3} \cmidrule(lr){4-6}
\textbf{Method} & \textbf{MATH} & \textbf{GSM8K} & \textbf{ARC-e} & \textbf{ARC-c} & \textbf{SIQA} & \textbf{Avg} \\
\midrule
\multicolumn{7}{l}{\textit{Baseline Methods}} \\
LoRA & 5.60 & 44.88 & 62.24 & 41.38 & \underline{45.29} & 39.88\\
PiSSA & 6.56 & 48.06 & 60.04 & 36.35 & 43.86 & 38.97\\
MiLoRA & 5.84 & 45.26 & \underline{67.12} & 41.21 & 44.92 & 40.87\\
LoRA-Null & 6.06 & 48.31 & 66.28 & 41.30 & 45.21 & 41.43\\
\midrule
\multicolumn{7}{l}{\textit{Our Methods}} \\
OPLoRA-16 & \textbf{7.04} & \textbf{49.73} & 66.88 & \underline{41.47} & 45.19 & \underline{42.06}\\
OPLoRA-128 & \underline{6.88} & \underline{49.35} & \
\textbf{67.26} & \textbf{42.24} & \textbf{45.50} & \textbf{42.25}\\
\bottomrule
\end{tabular}
\caption{Comparison of LoRA-based methods fine-tuned on MetaMathQA (first 100K samples) using LLaMA-2 7B. MATH and GSM8K are used to evaluate fine-tuning performance on in-domain tasks, while ARC-e, ARC-c, and SIQA assess the degree of forgetting on held-out reasoning benchmarks. Best and second-best scores are highlighted in bold and underlined.}
\label{tab:metamath_llama}
\end{table}

\begin{table}[t]
\centering
\small
\setlength{\tabcolsep}{2pt}
\begin{tabular}{lcc|ccc|c}
\toprule
\multicolumn{1}{c}{} & \multicolumn{2}{c}{\textbf{Fine-Tuning Eval}} & \multicolumn{3}{c}{\textbf{Forgetting Eval}} & \\
\cmidrule(lr){2-3} \cmidrule(lr){4-6}
\textbf{Method} & \textbf{MATH} & \textbf{GSM8K} & \textbf{ARC-e} & \textbf{ARC-c} & \textbf{SIQA} & \textbf{Avg} \\
\midrule
\multicolumn{7}{l}{\textit{Baseline Methods}} \\
LoRA & 45.96 & 82.18 & 81.40 & \underline{50.42} & 51.28 & 62.25\\
PiSSA & 32.74 & 74.45 & 72.47 & 38.65 & 44.37 & 52.54\\
MiLoRA & 44.04 & \underline{82.48} & 81.27 & 48.38 & 51.17 & 61.47\\
LoRA-Null & 43.72 & 75.12 & 73.22 & 49.89 & 46.44 & 57.68\\
\midrule
\multicolumn{7}{l}{\textit{Our Methods}} \\
OPLoRA-16 & \textbf{47.34} & \textbf{83.70} & \underline{81.86} & \textbf{50.59} & \underline{51.43} & \underline{62.98}\\
OPLoRA-128 &\underline{46.72} & \textbf{83.70} & \
\textbf{82.24} & 50.20 & \textbf{52.92} & \textbf{63.16}\\
\bottomrule
\end{tabular}
\caption{Comparison of LoRA-based methods fine-tuned on MetaMathQA (first 100K samples) using Qwen2.5 7B. Best and second-best scores are highlighted in bold and underlined.}
\label{tab:metamath_qwen}
\end{table}

\subsection{Python Code Generation}

To evaluate code generation capabilities, we fine-tune each model on the CodeFeedback dataset \cite{zheng2025opencodeinterpreterintegratingcodegeneration}. Performance is assessed using two in-domain benchmarks, MBPP and MBPP++, where we report functional correctness as measured by pass@1 accuracy—the percentage of generated programs that pass all unit tests on the first attempt. The results are presented in Table~\ref{tab:python_llama} and Table~\ref{tab:python_qwen} for LLaMA-2 7B and Qwen2.5 7B.

On LLaMA-2 7B, OPLoRA-16 achieves the highest performance on both MBPP (40.5\%) and MBPP++ (34.1\%), outperforming all baselines by a notable margin. OPLoRA-128 also performs strongly, achieving 39.2\% on MBPP and 31.0\% on MBPP++. The trends hold for Qwen2.5 7B as well. OPLoRA-16 obtains the top score on MBPP++ (69.0\%) and matches LoRA on MBPP (80.2\%), while OPLoRA-128 ranks second on MBPP (80.4\%) and MBPP++ (68.8\%). These findings confirm the flexibility of OPLoRA in handling both prompt-based and completion-style code generation benchmarks.

For forgetting, we evaluate the models on three held-out commonsense reasoning tasks: HellaSwag, OBQA, and SIQA. On LLaMA-2 7B, OPLoRA-128 achieves the highest accuracy on OBQA (45.6\%) and HellaSwag (77.0\%), while OPLoRA-16 obtains the best result on SIQA (47.1\%) and the second-best on HellaSwag. For Qwen2.5 7B, OPLoRA-128 delivers the highest forgetting resistance overall, achieving the best scores on HellaSwag (79.2\%), OBQA (47.4\%), and SIQA (55.5\%). OPLoRA-16 follows closely with strong performance across all three tasks.

\subsection{Subspace Interference Analysis}

We analyze subspace interference to better understand the mechanism of catastrophic forgetting in LoRA-based fine-tuning. Specifically, we quantify how much the learned LoRA update $\Delta W$ aligns with the top-$k$ singular directions of the frozen pre-trained weights $W_0$.

Figure~\ref{fig:forgetting_analysis}(a) presents $\rho_k$ values for $k \in \{8, 16, 32, 64, 128\}$ on the \texttt{q\_proj} layer across LoRA-based methods. We observe that PiSSA has significantly higher $\rho_k$, suggesting it heavily modifies pre-trained semantics and may cause knowledge overwriting. In contrast, OPLoRA-128 maintains consistently low alignment across all $k$, indicating that its updates stay mostly in orthogonal subspaces and are less likely to interfere with prior knowledge. While OPLoRA is designed to project out the top-$k$ directions explicitly, the $\rho_k$ values are not exactly zero due to the use of an efficient low-rank approximation via \texttt{svd\_lowrank}. Figure~\ref{fig:forgetting_analysis}(b) shows the average forgetting score across these benchmarks. A higher score indicates stronger retention of general knowledge. As shown, both OPLoRA variants outperform all baselines, achieving the best robustness to forgetting. In particular, OPLoRA-128 yields the highest average score, validating the hypothesis that subspace-aware updates are more effective at preserving pre-trained knowledge.

\begin{table}[t]
\centering
\small
\setlength{\tabcolsep}{2pt}
\begin{tabular}{lcc|ccc|c}
\toprule
\multicolumn{1}{c}{} & \multicolumn{2}{c}{\textbf{Fine-Tuning Eval}} & \multicolumn{3}{c}{\textbf{Forgetting Eval}} & \\
\cmidrule(lr){2-3} \cmidrule(lr){4-6}
\textbf{Method} & \textbf{MBPP} & \textbf{MBPP++} & \textbf{HSwag} & \textbf{OBQA} & \textbf{SIQA} & \textbf{Avg} \\
\midrule
\multicolumn{7}{l}{\textit{Baseline Methods}} \\
LoRA & 37.6 & 30.4 & \underline{76.9} & \underline{45.0} & 46.5 & 47.28\\
PiSSA & \underline{39.7} & \underline{31.5} & 75.4 & 44.6 & 46.7 & 47.58\\
MiLoRA & 37.3 & 30.4 & \textbf{77.0} & 44.6 & \underline{47.0} & 47.26\\
LoRA-Null & 37.0 & 29.4 & 76.7 & 44.8 & 46.7 & 46.92\\
\midrule
\multicolumn{7}{l}{\textit{Our Methods}} \\
OPLoRA-16 & \textbf{40.5} & \textbf{34.1} & \underline{76.9} & 44.8 & \textbf{47.1} & \textbf{48.68}\\
OPLoRA-128 & 39.2 & 31.0 & \
\textbf{77.0} & \textbf{45.6} & \underline{47.0} & \underline{47.96}\\
\bottomrule
\end{tabular}
\caption{Performance comparison of LoRA-based methods fine-tuned on CodeFeedback dataset using LLaMA-2 7B. MBPP and MBPP++ evaluate in-domain code generation performance, while HSwag, OBQA, and SIQA are used to assess forgetting and generalization to out-of-domain commonsense reasoning tasks. Best and second-best results are highlighted in bold and underlined.}
\label{tab:python_llama}
\end{table}

\begin{table}[t]
\centering
\small
\setlength{\tabcolsep}{2pt}
\begin{tabular}{lcc|ccc|c}
\toprule
\multicolumn{1}{c}{} & \multicolumn{2}{c}{\textbf{Fine-Tuning Eval}} & \multicolumn{3}{c}{\textbf{Forgetting Eval}} & \\
\cmidrule(lr){2-3} \cmidrule(lr){4-6}
\textbf{Method} & \textbf{MBPP} & \textbf{MBPP++} & \textbf{HSwag} & \textbf{OBQA} & \textbf{SIQA} & \textbf{Avg} \\
\midrule
\multicolumn{7}{l}{\textit{Baseline Methods}} \\
LoRA & 80.2 & 68.3 & 79.1 & \textbf{47.4} & 54.4 & 65.88\\
PiSSA & 68.3 & 58.2 & 78.0 & 43.2 & 51.7 & 59.88\\
MiLoRA & \textbf{80.7} & 66.7 & 78.6 & \underline{46.2} & \underline{55.1} & 65.46\\
LoRA-Null & 79.6 & 65.7 & 78.0 & 46.0 & 54.8 & 64.82\\
\midrule
\multicolumn{7}{l}{\textit{Our Methods}} \\
OPLoRA-16 & 80.2 & \textbf{69.0} & \underline{79.1} & \textbf{47.4}
& 55.0 & \underline{66.14}\\
OPLoRA-128 & \underline{80.4} & \underline{68.8} & \textbf{79.2} & \textbf{47.4} & \textbf{55.5} & \textbf{66.26}\\
\bottomrule
\end{tabular}
\caption{Performance comparison of LoRA-based methods fine-tuned on CodeFeedback dataset using Qwen2.5 7B. Best and second-best results are highlighted in bold and underlined.}
\label{tab:python_qwen}
\end{table}

\section{Conclusion}

In this work, we introduce OPLoRA, a novel fine-tuning strategy that mitigates catastrophic forgetting by constraining LoRA updates to avoid interference with the dominant subspaces of pre-trained weights. Through a rigorous subspace alignment analysis, we demonstrate that previous LoRA-based methods often disrupt critical directions in the model's weight space, leading to degraded generalization. OPLoRA addresses this issue by projecting updates orthogonally to the top singular vectors of the frozen weight matrices, thereby preserving essential pre-trained knowledge.

Extensive experiments show that OPLoRA consistently achieves competitive or superior performance compared to other LoRA variants while offering significantly improved forgetting resistance. Our results highlight the importance of subspace-aware adaptation in parameter-efficient fine-tuning, paving the way for more robust deployment of language models in continual and task-specific settings. Due to computational resource constraints, we leave the exploration of varying projection ranks and the extension of OPLoRA to larger-scale models such as LLaMA-3 70B for future work.

\section{Supplementary}
\label{sec:supp}

\subsection{Experiment Details}

All experiments are conducted on a single NVIDIA RTX A6000 GPU. The software environment consists of PyTorch 2.4.0, CUDA 12.1, FlashAttention 2.7.4, Hugging Face Hub 0.30.2, PEFT 0.14.0, and Transformers 4.45.1. We employ DeepSpeed ZeRO-2 for memory-efficient training. Both backbone models, LLaMA-2 7B \cite{llama2} and Qwen2.5  7B \cite{qwen2.5}, are publicly available via the Hugging Face.

We primarily use the lm\_eval 0.4.9 for evaluating commonsense reasoning tasks and measuring forgetting across all held-out benchmarks. For code generation evaluation, we adopt the evalplus 0.3.1. For mathematics-related tasks, we utilize the evaluation scripts provided by the PiSSA \cite{pissa}.

\subsection{Computational Efficiency}
Although OPLoRA applies double-sided projections, it does not require forming large dense projection matrices. Specifically, $\Delta Wx = P_L B A P_Rx=(I-U_kU_k^\top)BA(I-V_kV_k^\top)x$ is computed sequentially: $x \leftarrow x - V_k(V_k^\top x)$; $x \leftarrow BAx$; $x \leftarrow x - U_k(U_k^\top x)$.  All steps are low-rank. Let $B \in \mathbb{R}^{m \times r}$ and $A \in \mathbb{R}^{r \times n}$. Standard LoRA requires $2r(m + n)$ FLOPs per layer, while OPLoRA requires approximately $(4k + 2r + 1)(m + n)$ FLOPs per layer. In practice, the overhead is modest: on LLaMA-2 7B (Commonsense task), training with LoRA takes 5 h 13 m, whereas OPLoRA takes 6 h 12 m. This additional cost is small relative to overall training time, and scaling OPLoRA to 70B models is computationally feasible; our experiments are limited primarily by available compute resources.

\subsection{Low-Rank SVD}
OPLoRA relies on the top-$k$ singular triples of each weight matrix, which we obtain using a truncated SVD that directly extracts the leading singular vectors without a full decomposition. The one-time SVD initialization takes 5.5 minutes, which is negligible compared to the 6 h 12 m training time. Because truncated SVD is an iterative approximation, a small amount of leakage into the preserved subspace may occur. However, the resulting alignment metric $\rho_k$ remains extremely small in practice (Figure~\ref{fig:forgetting_analysis}(a)): for OPLoRA-128, $\rho_{128}$ is 0.003, and $\rho_{64}$ and all smaller $k$ are below $1.4 \times 10^{-5}$. These values confirm that our theoretical preservation guarantee is effectively maintained in practice.

\begin{table}[t]
\small
\centering
\begin{tabular}{lccc}
\toprule
\textbf{Hyperparameter} & \textbf{Commonsense} & \textbf{Math} & \textbf{Python} \\
\midrule
LoRA rank ($r$) & 32 & 64 & 32 \\
Scaling factor ($\alpha$) & 32 & 64 & 32 \\
LoRA dropout & \multicolumn{3}{c}{0.05} \\
Learning rate & 2e-4 & 3e-4 & 2e-4 \\
Batch size& \multicolumn{3}{c}{32} \\\
Epochs & \multicolumn{3}{c}{1} \\
Weight decay & \multicolumn{3}{c}{0.0} \\
Learning rate scheduler & \multicolumn{3}{c}{Cosine} \\
Warmup ratio & \multicolumn{3}{c}{0.03} \\
Precision & \multicolumn{3}{c}{bfloat16} \\
\bottomrule
\end{tabular}
\caption{Hyperparameter configuration used across all experiments.}
\label{tab:hyperparams}
\end{table}

%

\bibliography{aaai2026}

\begin{thebibliography}{37}
\providecommand{\natexlab}[1]{#1}

\bibitem[{Aghajanyan, Zettlemoyer, and Gupta(2020)}]{aghajanyan2020intrinsic}
Aghajanyan, A.; Zettlemoyer, L.; and Gupta, S. 2020.
\newblock Intrinsic dimensionality explains the effectiveness of language model fine-tuning.
\newblock \emph{arXiv preprint arXiv:2012.13255}.

\bibitem[{Amini et~al.(2019)Amini, Gabriel, Lin, Koncel-Kedziorski, Choi, and Hajishirzi}]{MathQA}
Amini, A.; Gabriel, S.; Lin, P.; Koncel-Kedziorski, R.; Choi, Y.; and Hajishirzi, H. 2019.
\newblock MathQA: Towards Interpretable Math Word Problem Solving with Operation-Based Formalisms.
\newblock arXiv:1905.13319.

\bibitem[{Austin et~al.(2021)Austin, Odena, Nye, Bosma, Michalewski, Dohan, Jiang, Cai, Terry, Le et~al.}]{MBPP}
Austin, J.; Odena, A.; Nye, M.; Bosma, M.; Michalewski, H.; Dohan, D.; Jiang, E.; Cai, C.; Terry, M.; Le, Q.; et~al. 2021.
\newblock Program synthesis with large language models.
\newblock \emph{arXiv preprint arXiv:2108.07732}.

\bibitem[{Biderman et~al.(2024)Biderman, Portes, Ortiz, Paul, Greengard, Jennings, King, Havens, Chiley, Frankle, Blakeney, and Cunningham}]{biderman2024loralearn}
Biderman, D.; Portes, J.; Ortiz, J. J.~G.; Paul, M.; Greengard, P.; Jennings, C.; King, D.; Havens, S.; Chiley, V.; Frankle, J.; Blakeney, C.; and Cunningham, J.~P. 2024.
\newblock Lo{RA} Learns Less and Forgets Less.
\newblock \emph{Transactions on Machine Learning Research}.
\newblock Featured Certification.

\bibitem[{Bisk et~al.(2020)Bisk, Zellers, Gao, Choi et~al.}]{piqa}
Bisk, Y.; Zellers, R.; Gao, J.; Choi, Y.; et~al. 2020.
\newblock Piqa: Reasoning about physical commonsense in natural language.
\newblock In \emph{Proceedings of the AAAI conference on artificial intelligence}, volume~34, 7432--7439.

\bibitem[{B{\"u}y{\"u}kaky{\"u}z(2024)}]{olora}
B{\"u}y{\"u}kaky{\"u}z, K. 2024.
\newblock Olora: Orthonormal low-rank adaptation of large language models.
\newblock \emph{arXiv preprint arXiv:2406.01775}.

\bibitem[{Clark et~al.(2019)Clark, Lee, Chang, Kwiatkowski, Collins, and Toutanova}]{boolq}
Clark, C.; Lee, K.; Chang, M.-W.; Kwiatkowski, T.; Collins, M.; and Toutanova, K. 2019.
\newblock {B}ool{Q}: Exploring the Surprising Difficulty of Natural Yes/No Questions.
\newblock In Burstein, J.; Doran, C.; and Solorio, T., eds., \emph{Proceedings of the 2019 Conference of the North {A}merican Chapter of the Association for Computational Linguistics: Human Language Technologies, Volume 1 (Long and Short Papers)}, 2924--2936. Minneapolis, Minnesota: Association for Computational Linguistics.

\bibitem[{Clark et~al.(2018)Clark, Cowhey, Etzioni, Khot, Sabharwal, Schoenick, and Tafjord}]{arc}
Clark, P.; Cowhey, I.; Etzioni, O.; Khot, T.; Sabharwal, A.; Schoenick, C.; and Tafjord, O. 2018.
\newblock Think you have solved question answering? try arc, the ai2 reasoning challenge.
\newblock \emph{arXiv preprint arXiv:1803.05457}.

\bibitem[{Cobbe et~al.(2021)Cobbe, Kosaraju, Bavarian, Chen, Jun, Kaiser, Plappert, Tworek, Hilton, Nakano, Hesse, and Schulman}]{gsm8k}
Cobbe, K.; Kosaraju, V.; Bavarian, M.; Chen, M.; Jun, H.; Kaiser, L.; Plappert, M.; Tworek, J.; Hilton, J.; Nakano, R.; Hesse, C.; and Schulman, J. 2021.
\newblock Training Verifiers to Solve Math Word Problems.
\newblock arXiv:2110.14168.

\bibitem[{Dou et~al.(2024)Dou, Zhou, Liu, Gao, Zhao, Shen, Zhou, Xi, Wang, Fan, Pu, Zhu, Zheng, Gui, Zhang, and Huang}]{dou2024loramoealleviateworldknowledge}
Dou, S.; Zhou, E.; Liu, Y.; Gao, S.; Zhao, J.; Shen, W.; Zhou, Y.; Xi, Z.; Wang, X.; Fan, X.; Pu, S.; Zhu, J.; Zheng, R.; Gui, T.; Zhang, Q.; and Huang, X. 2024.
\newblock LoRAMoE: Alleviate World Knowledge Forgetting in Large Language Models via MoE-Style Plugin.
\newblock arXiv:2312.09979.

\bibitem[{Farajtabar et~al.(2020)Farajtabar, Azizan, Mott, and Li}]{ogd}
Farajtabar, M.; Azizan, N.; Mott, A.; and Li, A. 2020.
\newblock Orthogonal gradient descent for continual learning.
\newblock In \emph{International conference on artificial intelligence and statistics}, 3762--3773. PMLR.

\bibitem[{He et~al.(2015)He, Zhang, Ren, and Sun}]{he2015delving}
He, K.; Zhang, X.; Ren, S.; and Sun, J. 2015.
\newblock Delving deep into rectifiers: Surpassing human-level performance on imagenet classification.
\newblock In \emph{Proceedings of the IEEE international conference on computer vision}, 1026--1034.

\bibitem[{Hendrycks et~al.(2021)Hendrycks, Burns, Kadavath, Arora, Basart, Tang, Song, and Steinhardt}]{math}
Hendrycks, D.; Burns, C.; Kadavath, S.; Arora, A.; Basart, S.; Tang, E.; Song, D.; and Steinhardt, J. 2021.
\newblock Measuring mathematical problem solving with the math dataset.
\newblock \emph{arXiv preprint arXiv:2103.03874}.

\bibitem[{Houlsby et~al.(2019)Houlsby, Giurgiu, Jastrzebski, Morrone, De~Laroussilhe, Gesmundo, Attariyan, and Gelly}]{pmlr-v97-houlsby19a}
Houlsby, N.; Giurgiu, A.; Jastrzebski, S.; Morrone, B.; De~Laroussilhe, Q.; Gesmundo, A.; Attariyan, M.; and Gelly, S. 2019.
\newblock Parameter-Efficient Transfer Learning for {NLP}.
\newblock In Chaudhuri, K.; and Salakhutdinov, R., eds., \emph{Proceedings of the 36th International Conference on Machine Learning}, volume~97 of \emph{Proceedings of Machine Learning Research}, 2790--2799. PMLR.

\bibitem[{Hu et~al.(2022)Hu, Shen, Wallis, Allen-Zhu, Li, Wang, Wang, Chen et~al.}]{lora}
Hu, E.~J.; Shen, Y.; Wallis, P.; Allen-Zhu, Z.; Li, Y.; Wang, S.; Wang, L.; Chen, W.; et~al. 2022.
\newblock Lora: Low-rank adaptation of large language models.
\newblock \emph{ICLR}, 1(2): 3.

\bibitem[{Hu et~al.(2023)Hu, Wang, Lan, Xu, Lim, Bing, Xu, Poria, and Lee}]{hu2023llm}
Hu, Z.; Wang, L.; Lan, Y.; Xu, W.; Lim, E.-P.; Bing, L.; Xu, X.; Poria, S.; and Lee, R. K.-W. 2023.
\newblock Llm-adapters: An adapter family for parameter-efficient fine-tuning of large language models.
\newblock \emph{arXiv preprint arXiv:2304.01933}.

\bibitem[{Kirkpatrick et~al.(2017)Kirkpatrick, Pascanu, Rabinowitz, Veness, Desjardins, Rusu, Milan, Quan, Ramalho, Grabska-Barwinska et~al.}]{ewc}
Kirkpatrick, J.; Pascanu, R.; Rabinowitz, N.; Veness, J.; Desjardins, G.; Rusu, A.~A.; Milan, K.; Quan, J.; Ramalho, T.; Grabska-Barwinska, A.; et~al. 2017.
\newblock Overcoming catastrophic forgetting in neural networks.
\newblock \emph{Proceedings of the national academy of sciences}, 114(13): 3521--3526.

\bibitem[{Lai et~al.(2017)Lai, Xie, Liu, Yang, and Hovy}]{RACE}
Lai, G.; Xie, Q.; Liu, H.; Yang, Y.; and Hovy, E. 2017.
\newblock {RACE}: Large-scale {R}e{A}ding Comprehension Dataset From Examinations.
\newblock In Palmer, M.; Hwa, R.; and Riedel, S., eds., \emph{Proceedings of the 2017 Conference on Empirical Methods in Natural Language Processing}, 785--794. Copenhagen, Denmark: Association for Computational Linguistics.

\bibitem[{Lester, Al-Rfou, and Constant(2021)}]{lester-etal-2021-power}
Lester, B.; Al-Rfou, R.; and Constant, N. 2021.
\newblock The Power of Scale for Parameter-Efficient Prompt Tuning.
\newblock In Moens, M.-F.; Huang, X.; Specia, L.; and Yih, S. W.-t., eds., \emph{Proceedings of the 2021 Conference on Empirical Methods in Natural Language Processing}, 3045--3059. Online and Punta Cana, Dominican Republic: Association for Computational Linguistics.

\bibitem[{Li et~al.(2018)Li, Farkhoor, Liu, and Yosinski}]{li2018measuringintrinsicdimensionobjective}
Li, C.; Farkhoor, H.; Liu, R.; and Yosinski, J. 2018.
\newblock Measuring the Intrinsic Dimension of Objective Landscapes.
\newblock arXiv:1804.08838.

\bibitem[{Li and Hoiem(2017)}]{lwf}
Li, Z.; and Hoiem, D. 2017.
\newblock Learning without forgetting.
\newblock \emph{IEEE transactions on pattern analysis and machine intelligence}, 40(12): 2935--2947.

\bibitem[{Liu et~al.(2024)Liu, Wang, Yin, Molchanov, Wang, Cheng, and Chen}]{dora}
Liu, S.-Y.; Wang, C.-Y.; Yin, H.; Molchanov, P.; Wang, Y.-C.~F.; Cheng, K.-T.; and Chen, M.-H. 2024.
\newblock Dora: Weight-decomposed low-rank adaptation.
\newblock In \emph{Forty-first International Conference on Machine Learning}.

\bibitem[{Liu et~al.(2022)Liu, Ji, Fu, Tam, Du, Yang, and Tang}]{liu2022ptuningv2prompttuning}
Liu, X.; Ji, K.; Fu, Y.; Tam, W.~L.; Du, Z.; Yang, Z.; and Tang, J. 2022.
\newblock P-Tuning v2: Prompt Tuning Can Be Comparable to Fine-tuning Universally Across Scales and Tasks.
\newblock arXiv:2110.07602.

\bibitem[{Meng, Wang, and Zhang(2024)}]{pissa}
Meng, F.; Wang, Z.; and Zhang, M. 2024.
\newblock Pissa: Principal singular values and singular vectors adaptation of large language models.
\newblock \emph{Advances in Neural Information Processing Systems}, 37: 121038--121072.

\bibitem[{Mihaylov et~al.(2018)Mihaylov, Clark, Khot, and Sabharwal}]{obqa}
Mihaylov, T.; Clark, P.; Khot, T.; and Sabharwal, A. 2018.
\newblock Can a suit of armor conduct electricity? a new dataset for open book question answering.
\newblock \emph{arXiv preprint arXiv:1809.02789}.

\bibitem[{Ramasesh, Dyer, and Raghu(2020)}]{ramasesh2020anatomy}
Ramasesh, V.~V.; Dyer, E.; and Raghu, M. 2020.
\newblock Anatomy of catastrophic forgetting: Hidden representations and task semantics.
\newblock \emph{arXiv preprint arXiv:2007.07400}.

\bibitem[{Riemer et~al.(2018)Riemer, Cases, Ajemian, Liu, Rish, Tu, and Tesauro}]{mer}
Riemer, M.; Cases, I.; Ajemian, R.; Liu, M.; Rish, I.; Tu, Y.; and Tesauro, G. 2018.
\newblock Learning to learn without forgetting by maximizing transfer and minimizing interference.
\newblock \emph{arXiv preprint arXiv:1810.11910}.

\bibitem[{Sakaguchi et~al.(2021)Sakaguchi, Bras, Bhagavatula, and Choi}]{winogrande}
Sakaguchi, K.; Bras, R.~L.; Bhagavatula, C.; and Choi, Y. 2021.
\newblock Winogrande: An adversarial winograd schema challenge at scale.
\newblock \emph{Communications of the ACM}, 64(9): 99--106.

\bibitem[{Sap et~al.(2019)Sap, Rashkin, Chen, LeBras, and Choi}]{siqa}
Sap, M.; Rashkin, H.; Chen, D.; LeBras, R.; and Choi, Y. 2019.
\newblock Socialiqa: Commonsense reasoning about social interactions.
\newblock \emph{arXiv preprint arXiv:1904.09728}.

\bibitem[{Tang et~al.(2025)Tang, Liu, Zhang, Wu, and Zhang}]{loranull}
Tang, P.; Liu, Y.; Zhang, D.; Wu, X.; and Zhang, D. 2025.
\newblock Lora-null: Low-rank adaptation via null space for large language models.
\newblock \emph{arXiv preprint arXiv:2503.02659}.

\bibitem[{Touvron et~al.(2023)Touvron, Martin, Stone, Albert, Almahairi, Babaei, Bashlykov, Batra, Bhargava, Bhosale et~al.}]{llama2}
Touvron, H.; Martin, L.; Stone, K.; Albert, P.; Almahairi, A.; Babaei, Y.; Bashlykov, N.; Batra, S.; Bhargava, P.; Bhosale, S.; et~al. 2023.
\newblock Llama 2: Open foundation and fine-tuned chat models.
\newblock \emph{arXiv preprint arXiv:2307.09288}.

\bibitem[{Wang et~al.(2024)Wang, Li, Wang, Chen, and Chen}]{milora}
Wang, H.; Li, Y.; Wang, S.; Chen, G.; and Chen, Y. 2024.
\newblock Milora: Harnessing minor singular components for parameter-efficient llm finetuning.
\newblock \emph{arXiv preprint arXiv:2406.09044}.

\bibitem[{Yang et~al.(2025{\natexlab{a}})Yang, Yang, Zhang, Hui, Zheng, Yu, Li, Liu, Huang, Wei, Lin, Yang, Tu, Zhang, Yang, Yang, Zhou, Lin, Dang, Lu, Bao, Yang, Yu, Li, Xue, Zhang, Zhu, Men, Lin, Li, Tang, Xia, Ren, Ren, Fan, Su, Zhang, Wan, Liu, Cui, Zhang, and Qiu}]{qwen2.5}
Yang, A.; Yang, B.; Zhang, B.; Hui, B.; Zheng, B.; Yu, B.; Li, C.; Liu, D.; Huang, F.; Wei, H.; Lin, H.; Yang, J.; Tu, J.; Zhang, J.; Yang, J.; Yang, J.; Zhou, J.; Lin, J.; Dang, K.; Lu, K.; Bao, K.; Yang, K.; Yu, L.; Li, M.; Xue, M.; Zhang, P.; Zhu, Q.; Men, R.; Lin, R.; Li, T.; Tang, T.; Xia, T.; Ren, X.; Ren, X.; Fan, Y.; Su, Y.; Zhang, Y.; Wan, Y.; Liu, Y.; Cui, Z.; Zhang, Z.; and Qiu, Z. 2025{\natexlab{a}}.
\newblock Qwen2.5 Technical Report.
\newblock arXiv:2412.15115.

\bibitem[{Yang et~al.(2025{\natexlab{b}})Yang, Li, Zhou, Song, Wu, Nie, and Ghanem}]{yang2025cordacontextorienteddecompositionadaptation}
Yang, Y.; Li, X.; Zhou, Z.; Song, S.~L.; Wu, J.; Nie, L.; and Ghanem, B. 2025{\natexlab{b}}.
\newblock CorDA: Context-Oriented Decomposition Adaptation of Large Language Models for Task-Aware Parameter-Efficient Fine-tuning.
\newblock arXiv:2406.05223.

\bibitem[{Yu et~al.(2023)Yu, Jiang, Shi, Yu, Liu, Zhang, Kwok, Li, Weller, and Liu}]{metamath}
Yu, L.; Jiang, W.; Shi, H.; Yu, J.; Liu, Z.; Zhang, Y.; Kwok, J.~T.; Li, Z.; Weller, A.; and Liu, W. 2023.
\newblock Metamath: Bootstrap your own mathematical questions for large language models.
\newblock \emph{arXiv preprint arXiv:2309.12284}.

\bibitem[{Zellers et~al.(2019)Zellers, Holtzman, Bisk, Farhadi, and Choi}]{hellaswag}
Zellers, R.; Holtzman, A.; Bisk, Y.; Farhadi, A.; and Choi, Y. 2019.
\newblock Hellaswag: Can a machine really finish your sentence?
\newblock \emph{arXiv preprint arXiv:1905.07830}.

\bibitem[{Zheng et~al.(2025)Zheng, Zhang, Shen, Liu, Lin, Fu, Chen, and Yue}]{zheng2025opencodeinterpreterintegratingcodegeneration}
Zheng, T.; Zhang, G.; Shen, T.; Liu, X.; Lin, B.~Y.; Fu, J.; Chen, W.; and Yue, X. 2025.
\newblock OpenCodeInterpreter: Integrating Code Generation with Execution and Refinement.
\newblock arXiv:2402.14658.

\end{thebibliography}

\end{document}